\documentclass{llncs}
\RequirePackage{fix-cm}
\usepackage[utf8]{inputenc}
\usepackage{graphicx}
\usepackage{amsmath}
\usepackage{amssymb}
\usepackage{amsfonts, mathrsfs}
\usepackage{mathtools}
\usepackage{multirow}
\usepackage{stmaryrd}
\usepackage{multirow}
\usepackage{etoolbox}

\usepackage[caption=false]{subfig}
\usepackage[hyphens]{url}
\usepackage[breaklinks]{hyperref}
\usepackage{booktabs} 
\usepackage[bottom]{footmisc}

\def\ivBrack#1{\left\llbracket{#1}\right\rrbracket} 
\def\FWER#1{FWER}
\def\fdr{\mathrm{FDR}}
\def\fnr{\mathrm{FNR}}
\def\fOne{F_{1}}
\def\mcc{\mathrm{MCC}}

\def\vec#1{{#1}}
\def\featSpace{\vec{\mathcal{X}}}

\def\orcidID#1{\unskip$^{[#1]}$}

\begin{document}

\frontmatter          
\pagestyle{headings}  
\addtocmark{} 
\mainmatter              

\title{Building an Ensemble of  Classifiers via Randomized Models of Ensemble Members}

\author{Pawel Trajdos\orcidID{0000-0002-4337-6847}\inst{1} \and Marek Kurzynski\orcidID{0000-0002-0401-2725}\inst{1}}
\authorrunning{Pawel Trajdos \and Marek Kurzynski} 
\tocauthor{Pawel Trajdos, Marek Kurzynski}

\institute{Department of Systems and Computer Networks,\\ 
    Faculty of Electronics, Wrocław University of Science and Technology,\\
    Wybrzeże Wyspiańskiego 27, 50-370 Wrocław, Poland\\
\email{pawel.trajdos@pwr.edu.pl}, \email{marek.kurzynski@pwr.edu.pl}}

\maketitle              

\begin{abstract}
 
Many dynamic ensemble selection (DES) methods are known in the literature. A previously-developed by the authors, method consists in building a randomized classifier which is treated as a model of the base classifier. The model is equivalent to the base classifier in a certain probabilistic sense. Next, the probability of correct classification of randomized classifier is taken as  the competence of the evaluated classifier.

In this paper, a novel randomized model of base classifier is developed. In the proposed method, the random operation of the model results from a random selection of the learning set from the family of learning sets of a fixed size. The paper presents the mathematical foundations of this approach and shows how, for a practical application  when learning and validation sets are given, one can determine the measure of competence and build a MC system with the DES scheme.

The DES scheme with the proposed model of competence was experimentally evaluated on the collection of 67 benchmark datasets and compared in terms of eight quality criteria with two ensemble classifiers which use the previously-proposed concepts of randomized model.
The proposed approach achieved the lowest ranks for almost all investigated quality criteria.


\end{abstract}

\section{Introduction}

In ensemble classification methods
the recognition result is obtained by combining the responses 
of ensemble  members. The literature on MC systems has grown rapidly in the last 30 years and includes a variety of proposed and developed concepts, algorithms and techniques of ensemble classification methods as well as a wide range of their practical applications.

Among the different methods of operation of the MC systems, the dynamic ensemble selection (DES) scheme seems to be the most mature and promising approach. In the DES methods, base classifiers are dynamically (individually for each recognized object) selected from the entire set (pool) and then classification results of selected classifiers are combined by weighted majority voting \cite{Ko2008}. In most methods, the base classifiers are selected from the pool according to their accuracy measure (competence) in a local region of the feature space. It means that only an ensemble containing the most competent classifiers is selected to predict the label of a specific test sample. 
This approach is fully justified because usually, there are no classifiers in the pool that would be experts for all recognized objects, rather base classifiers are expert in different local regions.

In \cite{Britto2014,Cruz2018}, we can find a comprehensive review of different DES methods containing the taxonomy and detailed characteristics of DES schemes and their comparative experimental analysis.

Although the presented methods differ in algorithms for determining classifier competence and in the interpretation of competence models justifying the division of methods into different categories (ranking-based, accuracy-based, probabilistic-based and behavior-based measures), they are dominated by solutions in which competence at a point is determined based on the properties of the tested classifier in terms of correct/incorrect classification of validation objects located in the neighborhood of the object being recognized. 

Such techniques seem to be a justified approach if we notice that there are serious problems with the natural measure of classifier competence for the feature vector $x$, that is the probability of correct classification at this point.
Firstly, this requires the adoption of a probabilistic model of the recognition task which in the case of many classifiers (e.g. artificial neural network etc.) is simply unjustified. Secondly, for deterministic base classifiers, the probability of correct classification at the point $x$ is equal to either 0 or 1. Thus, the probability of correct classification in no way differentiates (in a continuous scale) base classifiers, but divides them into classifiers correctly/incorrectly classifying the object $x$.

In this study, we explicitly adopt the probability of correct classification at point $x$ as a measure of classifier competence at this point, but we will do it indirectly through the randomized classifier being a probabilistic model of base classifier.
The randomized classifier makes a decision randomly, depending on the value of a certain random variable, so its probability of correct classification at a point $x$ is a number belonging to the $[0, 1]$ interval.
The randomized model is constructed in such a way that, in a certain probabilistic sense, it is equivalent to the base classifier. Therefore its probability of correct classification is taken as a point measure of the competence of the evaluated classifier. 

This approach has already been considered by the authors in earlier papers \cite{Woloszynski2011,Trajdos2019a}. The concept of the probabilistic model presented there leads to a randomized classifier whose values of classifying functions are the observed values of random variables with a probability distribution related to the values of classification functions produced by the evaluated base classifier.
In the original probabilistic model, called Randomized Reference Classifier (RRC), the beta distribution was used. In the modified model, we applied the truncated Gaussian distribution.
The developed concept for determining the competence of base classifiers through their randomized models proved to be a very effective approach.
This is evidenced by the results of comparative studies presented in the source publications and above all the results presented in the review paper containing the state-of-the-art in the field of MC systems with DES scheme \cite{Cruz2018}. In experimental comparative studies of 30 MC systems using 30 benchmark databases, the RRC method took 3rd place in terms of average rank and 5th place in terms of average accuracy.

In this paper, a novel randomized model of a base classifier is proposed in which the random operation of the model results from a random selection of the learning set from the family of learning sets of a fixed size. This approach means that the concept of the base classifier is its randomized model determining the algorithm for processing features in the context of a training set before this set is randomly selected.
 
Although randomization has been applied for a long time in decision-making systems (see e.g. \cite{deGroot1970,Berger1985}), there are not many solutions in the literature that use a random mechanism to build multiple classifier systems. First of all, Kleinberg's method of stochastic discrimination (SD) must be mentioned here \cite{Kle1990} and the random forest algorithm derived from SD. Noteworthy, there are bagging and boosting methods related to the construction of an ensemble of classifiers based on different training sets created in the random resampling process. In~\cite{SANTUCCI20171} interesting concept of building an ensemble of classifiers through randomization is presented, in which randomization procedure is associated with the random selection of the parameters of the base classifiers. 

The method presented in the paper follows the trend of using randomization procedures to build a multiclassifier system. However, this is done originally and differently from the techniques listed above.  
Randomization in the presented approach allows determining the measure of competence, based directly on the probability of correct classification (at a given point) which evaluates the quality of the base classifier on a continuous scale.

The paper is organized as follows. 
Section~\ref{sec:CompetenceExplained} introduces the formal model of classification problem and gives an insight into the original method of determining competence via randomized model of classifier. 
Section~\ref{sec:ProposedCompetenceModel} contains a description of the proposed method.
Section~\ref{sec:DesSchemeSystem} presents the multiclassifer system with DES scheme using the proposed method of determining competence measure.
Section~\ref{sec:Experiments} provides results of experimental comparative analysis.
Section~\ref{sec:Conclusions} concludes the paper.

\section{Competence of Base Classifier}\label{sec:CompetenceExplained}

\subsection{Preliminaries}\label{sec:CompetenceExplained:preliminaries}

Let us consider a pattern recognition problem in which $x \in \mathcal{X}=\Re^d$ denotes feature vector of an object and  $j \in \mathcal{M}=\{1,2,3,...,M\}$. is its class number. 
Let $\psi$ be a trained classifier which maps the feature space $\mathcal{X}$ into the set of class numbers $\mathcal{M}$, viz.:
\begin{align}  \label{wzor1}
\psi(x)=i,\ \   x \in \mathcal{X}, \  i \in \mathcal{M}.
\end{align}
We assume that $\psi$ is described by the canonical model \cite{Kuncheva2014combining}, i.e. for a given $x$, it first produces values of normalized classification functions (supports) $g_i(x), i \in \mathcal{M}$ ($g_i(x) \in [0, 1], \sum g_i(x)=1$) and then classifies an object according to the maximum support rule:
\begin{align}   \label{wzor2}
\psi(x)=i  \Leftrightarrow  g_i(x) = \max_{k \in \mathcal{M}} g_k(x).
\end{align}


Let $c(\psi|x)$ be a competence measure of classifier $\psi$ at the point $x \in \mathcal{X}$.
There exist several methods of building  local competence measure $c(\psi|x)$ based on different paradigms. Their exhaustive review can be found in \cite{Cruz2018}. 

\subsection{Competence via Randomization}\label{sec:CompetenceExplained:competenceviarandomization}

It seems that the most natural competence measure of the classifier $\psi$ for the feature vector $x$ is the probability of correct classification $P_c(\psi(x))$.
If $\psi$ is a deterministic classifier, its operation  at point $x$ is strictly dichotomous: classifier correctly classifies $x$ belonging to the $j$-th class if $\psi(x)=j$, or misclassifies $x$ if $\psi(x) \neq j$. It means that probability of correct classification $Pc(\psi(x))$ of classifier $\psi$ at the point $x$ is either $0$ or $1$, viz.: 
\begin{align} \label{wzor3}
   Pc(\psi(x)) &= \ivBrack{\psi(x)=j},
\end{align}
where $\ivBrack{\cdot}$ denotes the Iverson bracket.

Therefore, a direct application of $Pc(\psi(x))$ as a competence measure $c(\psi|x)$ is not used in this paper. 

In the proposed method, we first built a probabilistic model $\bar{\psi}$  of classifier $\psi$. The model $\bar{\psi}$ depends on random variable $Y$ with probability distribution $P(Y)$. It means that  the model $\bar{\psi}(Y)$ is the randomized classifier which makes classification randomly according to a probability distribution $P(\bar{\psi}(Y))$ dependent on $P(Y)$ and $\bar{\psi}$ as a function of $Y$ \cite{Berger1985}.
Formally, $\bar{\psi}$ is also a function from $\mathcal{X}$ into $\mathcal{M}$ specifying how decisions are made given observation $y$ of the random variable $Y$. 
It is obvious that distribution $P(Y)$ and construction of $\bar{\psi}$ should be proposed in such a way that the model is equivalent to $\psi$ in a certain probabilistic sense. 

The probability of correct classification of $P_c(\bar{\psi}(x))$ at point $x$ belongs to the interval $[0, 1]$. Therefore, we can use the probability of the correct classification of the randomized classifier as a measure of the competence of the classifier $\psi$ at point $x$, i.e.:
\begin{align}  \label{wzor4}
c(\psi|x) = P_c(\bar{\psi}(x)).
\end{align} 
The key problem of the proposed approach is related to the definition of probability distribution $P(Y)$ which results from the adopted relationship between the classifier $\psi$ and its model $\bar{\psi}$.

\subsubsection{Randomized Reference Classifier}

RRC is a probabilistic classifier which classifying functions $\{\delta_j(x)\}_{j \in \mathcal{M}}$ are observed values of multidimensional random variable $Y = \{\Delta_j(x)\}_{j \in \mathcal{M}}$ fulfilling the following conditions: (i) $\Delta_{i}(x) \in [0,1]$, (ii) $\sum_{i \in \mathcal{M}}\Delta_{i}(x) = 1$, (iii) $\mathbf{E}\left[\Delta_{i}(x) \right] = g_{i}(x)$,
where $\mathbf{E}$ is the expected value operator.  Conditions (i)  and (ii)  arise from the normalization properties of class supports, whereas condition (iii) provides the equivalence of the randomized model $\bar{\psi}_{RRC}$ and base classifier $\psi$. Based on the latter condition, the RRC can be used to provide randomized model of any classifier that returns vector of class-specific supports $g_i(x), i \in \mathcal{M}$. 

It is obvious that the probability of correct classification of an object $x$ belonging to the class  $j$ using the RRC is as follows:
\begin{align}   \label{wzor8}
 P_c(\bar{\psi}_{RRC}(x))=P [\Delta_j(x) > \Delta_k(x), k \in \mathcal{M} \setminus j].
\end{align}
The probability on the right side of \eqref{wzor8} can be easily determined if we assume -- as in the original work \cite{Woloszynski2011} -- that  random variables $\Delta_i(x)$ have the beta distribution.

Since $\bar{\psi}_{RRC}$ acts -- on average -- as the modeled base classifier $\psi$, the formula \eqref{wzor4} is fully justified. 

\subsubsection{RRC with Gaussian Distribution}

In the paper \cite{Trajdos2019a}, we proposed the use of Gaussian distribution truncated to the inverval $[0,1]$  instead of the beta distribution suggested in the original RRC model. It means that in the proposed randomized model $\bar{\psi}_{GD}$ random variable $Y$ is defined as in the RRC model. The expected value for each random variable $\Delta_i(x)$  is simply determined using formula (iii), but the standard deviation is calculated using a rescaled variance of the beta distribution with parameter that should be experimentally tuned in order to achieve the best classification accuracy.

\section{The Proposed Model of Competence}\label{sec:ProposedCompetenceModel}

\subsection{The Mathematical Fundamentals}\label{sec:ProposedCompetenceModel:mathFundamentals}

We suppose that classifier $\psi$ is trained in the supervised learning procedure using a learning set
\begin{align}  \label{wzor10}
s_n=\{ (x_1,j_1), (x_2,j_2), \ldots (x_n,j_n) \},
\end{align}
containing $n$ learning objects and their true classes. 
Let $\mathscr{S}_n$ be a family of all possible sets of a fixed size $n$. 

Since the dependence of the classifier on the learning set is important for the proposed competence model, we will now treat classifier $\psi$ as a function of two variables:
\begin{align}  \label{wzor11}
\psi(x,s_n)=i,\ \   x \in \mathcal{X}, \ s_n \in \mathscr{S}_n, \ i \in \mathcal{M}.
\end{align}

We assume that over the family $\mathscr{S}_n$, random variable $S_n$ with probability distribution $P(S_n)$ is defined.  Learning set $s_n$ is the observed value of $S_n$. It means that set $s_n$ is randomly selected from the family $\mathscr{S}_n$ according to  $P(S_n)$.
In the proposed method, we adopt $Y=S_n$.  As a consequence, we get the model $\bar{\psi}=\psi(S_n)$.

The difference between $\psi(x,s_n)$ and its model $\psi(x,S_n)$ is obvious.
$\psi(x,S_n)$ is a random variable presenting the concept of classifier construction or the algorithm of its activity describing how to process object features $x$ in the context of variables $S_n$ to obtain the classification result. 
In other words, $\psi(x,S_n)$  refers to the general classifier concept before the training set $s_n$ has been drawn from $\mathscr{S}_n$.
Thus, $\psi(x,S_n)$ is randomized classifier whose classification depends on the result of the draw $s_n$ from the family $\mathscr{S}_n$.
In turn, $\psi(x,s_n)$ is a classifier trained using a randomly selected learning set $s_n$ which classifies object $x$ into the specific class $i$ from the set $\mathcal{M}$. 
As it results from the above description, in the proposed method the probability of correct classification of the concept $\psi(x,S_n)$ averaged over the family $\mathscr{S}_n$ is used as the competence of the classifier $\psi(x,s_n)$ being the realization of this concept.

Since random variable $ \ivBrack{\psi(x,S_n)=j }$ has binary (Bernoulli) distribution (see formula (\ref{wzor3})) with success and failure probability
equal to $P_c$ and $1-P_c=P_e$, respectively. We have the following formula for the probability of correct classification of the model $\bar{\psi}$ at the point $x$:
\begin{align}   \label{wzor12}
P_c(\bar{\psi}(x))&=P_c(\psi(x,S_n))=E_{S_n} \ivBrack{\psi(x,S_n)=j }\\
&= \int_{\mathscr{S}_n} \ivBrack{\psi(x,s_n)=j } \; dP(s_n).
\end{align}
If probability distribution of the draw of set $s_n$ from $\mathscr{S}_n$ is uniform, then we have 
\begin{align}   \label{wzor13}
P_c(\bar{\psi}(x))=P_c(\psi(x, S_n))=\frac{\left\| s_n: P_c(\psi(x, s_n))=1  \right\|}{\left\| \mathscr{S}_n  \right\|},
\end{align}
where ${\left\| A \right\|}$ denotes Lebesgue measure of $A$. 

\subsection{Empirical Case}

Suppose now that the finite set of learning sets of size $n$ is available:
\begin{align}   \label{wzor14}
(s_n^{(1)}, s_n^{(2)},..., s_n^{(K)}).
\end{align}
Then estimation of the average correct classification probability \eqref{wzor13} can be calculated as follows:
\begin{align}  \label{wzor15}
P_c(\psi(x, S_n)) \approx \frac{K_c}{K},
\end{align}
where $K_c$ denotes the number of learning sequences for which $P_c(\psi(x,s_n))=1$.

In this case, we have empirical competence measure of classifier $\psi$ at the point $x$ which is the estimated value of probability of correct classification of $\psi$ at this point averaged over learning sets of the fixed size $n$. 

\subsection {Competence Set}

To calculate \eqref{wzor15}, the true class of the object $x$ must be known. Thus, we assume that a validation set 
\begin{align}  \label{wzor16}
\mathcal{V}=\{ (x_1,j_1), (x_2,j_2), \ldots (x_N,j_N) \}.
\end{align}
containing pairs of feature vectors and their true class labels is available. 
In this case, from \eqref{wzor4}, \eqref{wzor15} and \eqref{wzor16} we have (in the feature space  $\featSpace$) the set of $N$ validation objects with competences of classifer $\psi$:
\begin{align}  \label{wzor17}
\mathcal{C}_{\psi}=\{(x_1, c(\psi |x_1)), (x_2, c(\psi |x_2)), \ldots (x_N, c(\psi |x_N))\}. 
\end{align}
Set $\mathcal{C}_{\psi}$ is called the competence set of classifier $\psi$. 

\subsection{Generalization Method -- Potential Function Model}

In the second step, based on the competence set \eqref{wzor17}, the competence function (measure) $c(\psi|x)$ is determined. In other words, information contained in the set $\mathcal{C}_{\psi}$, i.e. values of competence for validation points $x_k \in \mathcal{V}$, is generalized to the whole feature space $\featSpace$.
In the potential function method of generalization, the feature vectors $x_{k}$  are considered to be the locations of the competence sources $c(\psi|x_{k})$ that influence the entire feature space $\mathcal{X}$  creating a competence field. The competence at $x$ is a result of the cumulative influence of the sources, where the influence of each source is proportional to $c(\psi|x_{k})$ and it decreases as the distance between $x_k$ and $x$ increases. This interpretation  allows using the  normalized Gaussian potential function model \cite{Woloszynski2012} to construct the competence function as follows:
\begin{align}   \label{wzor18}
c(\psi|x)=\frac{1}{D}\sum_{x_{k} \in \mathcal{V}} c(\psi|x_{k}) \exp(-\mathrm{dist}(x_k,x)^2), 
\end{align}
where $D$ denotes the normalizing factor  and $\mathrm{dist}(x,y)$ is the Euclidean distance between $x$ and $y$.

\section{Multiclassifier System with DES Scheme}\label{sec:DesSchemeSystem}

In the multiclassifier system (MCS), a set of trained classifiers  $\Psi=\{\psi_1, \psi_2, \ldots, \psi_L\}$ called base classifiers is given. We assume that all classifiers from the set $\Psi$ meet the assumptions presented in subsections~\ref{sec:CompetenceExplained:preliminaries} and~\ref{sec:ProposedCompetenceModel:mathFundamentals}. For each classifier $\psi_l \in \Psi$, the competence function $c(\psi_l|x)$ is calculated according to \eqref{wzor4}, \eqref{wzor15} and \eqref{wzor18}. 

The proposed multiclassifier system uses a dynamic ensemble selection (DES) strategy which consists of two steps.
In the first step, an ensemble of competent classifiers is selected from the entire set $\Psi$ for a given $x$ 
\begin{align}  \label{wzor23}
\Psi(x)=\{\psi_{l1},\psi_{l2},...,\psi_{lx}: c(\psi_{lm}|x)>\alpha\},
\end{align}
where the threshold value $\alpha$ is arbitrary, but it is usually assumed that $\alpha=1/M$. This step eliminates inaccurate classifiers (worse than random guessing) and keeps the ensemble relatively diverse. 

The selected classifiers are combined using a weighted vector of class supports, where the weights are equal to the competences, viz.:
\begin{align}
\label{MK:wzor24}
g_{j}^{(MCS)}(x) = \sum _{\psi_l \in \Psi(x)}c(\psi_l|x)\; g_{j}^{(\psi_l)}(x).
\end{align}
Finally, the MCS system  classifies object $x$  using the maximum rule:
\begin{align}    \label{MK:wzor25}
\psi^{(MCS)}(x)=i\;\Leftrightarrow \; g_{i}^{(MCS)}(x)=\max_{j \in \mathcal{M}} g_j^{(MCS)}(x).
\end{align}

%
It should be emphasized that the computationally complex procedures for determining the source competence measure for validation objects -- thanks to the concept of the competence set -- is fully implemented in the learning phase.
In the classification phase, only a simple (and quick) procedure of determining the classifying functions of the ensemble classifier and making decisions according to the maximum principle remains to be implemented.

\section{Experiments}\label{sec:Experiments}
The experimental study is aimed at comparing the proposed method of classifier selection with the original approach proposed in \cite{Woloszynski2011}, and its modification using the Gaussian Distribution \cite{Trajdos2019a}. This paper does not contain a comparison between RRC based DES algorithms and the state-of-the-art methods of building classifier ensembles. This is because that kind of comparison has already been done \cite{Cruz2018}.
\subsection{Experimental Setup}

During the experimental evaluation the following heterogeneous ensemble classifiers  were compared:  $\psi_{\mathrm{C}}$ -- DES scheme with the proposed model of competence, $\psi_{\mathrm{B}}$ -- DES scheme proposed in \cite{Woloszynski2011},   $\psi_{\mathrm{N}}$ -- DES scheme proposed in \cite{Trajdos2019a}.

The following base classifiers were used to constitute each of the ensembles: $\psi_{\mathrm{NB}}$ -- Naive Bayes classifier with kernel density estimation, $\psi_{\mathrm{J48}}$ -- Weka version of the C4.5 algorithm with  Laplace smoothing, $\psi_{\mathrm{NC}}$ -- Nearest centroid (Nearest Prototype),    $\psi_{\mathrm{SVM}}$ -- SVM classifier with linear kernel (no kernel),    $\psi_{\mathrm{LDA}}$ -- LDA classifier,  $\psi_{\mathrm{QDA}}$ -- QDA classifier,  $\psi_{\mathrm{KNN}}$ -- Nearest neighbour classifier.

Each of the above-mentioned base classifiers was trained using 31 bootstrap samples from the training dataset.  The outcomes of the classifiers built using bootstrap samples were used to calculate the correct classification probability using formula~\eqref{wzor15}. 

The experimental code was implemented using WEKA framework. The source code of the algorithms is available online~\footnote{\url{https://github.com/ptrajdos/rrcBasedClassifiers/tree/develop}}. 

To evaluate the proposed methods, the following classification-loss criteria were used \cite{Sokolova2009}: macro-averaged and micro-averaged $\fdr$ (1- precision), $\fnr$ (1-recall), $\fOne$, Matthews correlation  coefficient ($\mcc$). Macro and micro averaged criteria were used because those families of criteria differ significantly in scoring majority and minority classes. That is, macro averaged criteria are more sensitive to performance for the minority class. Whereas micro averaged criteria are more sensitive to majority class performance. 

Following the recommendations of \cite{demsar2006}, the statistical significance of the obtained results was assessed using the two-step procedure. The first step is to perform the Friedman test \cite{demsar2006} for each quality criterion separately. Since the multiple criteria were employed, the familywise errors (\FWER{}) should be controlled \cite{Bergmann1988}. To do so, the Bergman-Hommel \cite{Bergmann1988} procedure of controlling \FWER{} of the conducted Friedman tests was employed. When the Friedman test shows that there is a significant difference within the group of classifiers, the pairwise tests using the Wilcoxon signed-rank test \cite{demsar2006} were employed. To control \FWER{} of the Wilcoxon-testing procedure, the Bergman-Hommel approach was employed \cite{Bergmann1988}. For all tests the significance level was set to $\alpha=0.05$.

The experimental evaluation was conducted on the collection of the 67 benchmark datasets taken from the Keel repository. Benchmark datasets are available online~\footnote{\url{https://github.com/ptrajdos/MLResults/blob/master/data/KeelData.tar.xz}}.

During the preprocessing stage, the datasets underwent a~few transformations. First, all nominal attributes were converted into a~set of binary variables. The transformation is necessary whenever the distance-based algorithms are employed \cite{Tian2005}. To reduce the computational burden and remove irrelevant information, the PCA procedure with the variance threshold set to 95\% was applied. The features were also normalized to have zero mean value and unit variance.

\subsection{Results and Discussion}

To compare multiple algorithms on multiple benchmark sets, the average ranks approach \cite{demsar2006} is used. Due to the page limit, full results are published online~\footnote{\url{https://github.com/ptrajdos/MLResults/raw/a3b4168a0b0aabee7ef8cd1056baf4a8578a9f6d/RandomizedClassifiers/CORES2021.zip}}. .

The numerical results are given in Table~\ref{table:KNNStat}. The table is divided into eight sections -- one section is related to a single evaluation criterion. The first row of each section is the name of the quality criterion investigated in the section. The second row shows the p-value of the Friedman test. The third one shows the average ranks achieved by algorithms. The following rows show p-values resulting from pairwise Wilcoxon test. Above each section the names of the investigated algorithms are placed.

{
\begin{table}[ht]
\centering\scriptsize
\caption{Statistical evaluation. Wilcoxon test results.\label{table:KNNStat}}
\begin{tabular}{l @{\hskip 0.2in} ccc@{\hskip 0.2in} ccc@{\hskip 0.2in} ccc@{\hskip 0.2in} ccc}
  & $\psi_{\mathrm{C}}$ & $\psi_{\mathrm{B}}$ & $\psi_{\mathrm{N}}$ & $\psi_{\mathrm{C}}$ & $\psi_{\mathrm{B}}$ & $\psi_{\mathrm{N}}$ & $\psi_{\mathrm{C}}$ & $\psi_{\mathrm{B}}$ & $\psi_{\mathrm{N}}$ & $\psi_{\mathrm{C}}$ & $\psi_{\mathrm{B}}$ & $\psi_{\mathrm{N}}$ \\ 
  \cmidrule(lr){2-4} \cmidrule(lr){5-7} \cmidrule(lr){8-10} \cmidrule(lr){11-13}
Nam.&\multicolumn{3}{c}{MaFDR}&\multicolumn{3}{c}{MaFNR}&\multicolumn{3}{c}{MaF1}&\multicolumn{3}{c}{Macro Matthews}\\
Frd.&\multicolumn{3}{c}{1.000e+00}&\multicolumn{3}{c}{2.523e-01}&\multicolumn{3}{c}{3.470e-01}&\multicolumn{3}{c}{1.000e+00}\\
Rank & 1.955 & 2.061 & 1.985 & 1.795 & 1.992 & 2.212 & 1.803 & 2.000 & 2.197 & 1.939 & 1.932 & 2.129 \\ 
$\psi_{\mathrm{C}}$ &  & .457 & .457 &  & .174 & \textbf{.036} &  & .150 & \textbf{.007} &  & .236 & .224 \\ 
  $\psi_{\mathrm{B}}$ &  &  & .652 &  &  & .148 &  &  & .154 &  &  & .224 \\ 
   \cmidrule(lr){2-4} \cmidrule(lr){5-7} \cmidrule(lr){8-10} \cmidrule(lr){11-13}
Nam.&\multicolumn{3}{c}{MiFDR}&\multicolumn{3}{c}{MiFNR}&\multicolumn{3}{c}{MiF1}&\multicolumn{3}{c}{Micro.Matthews}\\
Frd.&\multicolumn{3}{c}{1.000e+00}&\multicolumn{3}{c}{1.000e+00}&\multicolumn{3}{c}{1.000e+00}&\multicolumn{3}{c}{1.000e+00}\\
Rank & 1.871 & 1.985 & 2.144 & 1.871 & 1.985 & 2.144 & 1.871 & 1.985 & 2.144 & 1.864 & 1.992 & 2.144 \\ 
$\psi_{\mathrm{C}}$ &  & .411 & .411 &  & .411 & .411 &  & .411 & .411 &  & .386 & .370 \\ 
  $\psi_{\mathrm{B}}$ &  &  & .411 &  &  & .411 &  &  & .411 &  &  & .370 \\ 
  \end{tabular}
\end{table}
}


According to the table, the proposed approach achieved the lowest ranks for all the investigated quality criteria except macro-averaged $\mcc$. For the macro-averaged $\mcc$, the proposed approach is the second one; however, the difference between the average ranks is very small. 

On the other hand, the approach based on the Gaussian distribution achieves the lowest rank in terms of all quality criteria except macro-averaged $\fdr$.

However, the conducted statistical tests show that there are almost no significant differences between the investigated algorithms. The significant differences are only between the proposed method and the Gaussian-based RRC-DES. These differences show that the proposed approach is better in terms of macro averaged $\fnr$ and macro-averaged $\fOne$ measure. It means that the proposed approach is better at discovering minority class objects. This is done without lowering precision.  This property may be useful for the imbalanced classification task. The results show that using the data-driven distribution for the RRC model may improve the classification quality.

\section{Conclusions}\label{sec:Conclusions}

In this paper, we proposed a data-driven approach to building the RRC-based dynamic-ensemble-selection classifier. The classifier was compared with the original RRC-based approach and the RRC-method built using  Gaussian distribution. The conducted experimental study shows that there are almost no significant differences between the proposed method and the previously-proposed algorithms. The only observed differences may suggest that using the proposed approach in imbalanced classification problems may allow a significant improvement over the previously mentioned methods. However, this result is achieved at the cost of training an ensemble of classifiers using bootstraped datasets. Consequently, future research is aimed at building a distribution-driven RRC approach with the reduced computational burden.

\section*{Acknowledgment} This work was supported by the statutory funds of the Department of Systems and Computer Networks, Wroclaw University of Science and Technology.

 \bibliography{bibliography}

\end{document}